\pdfoutput=1

\documentclass[11pt]{article}

\usepackage[final]{acl}

\usepackage{times}
\usepackage{latexsym}

\usepackage[T1]{fontenc}

\usepackage[utf8]{inputenc}

\usepackage{microtype}

\usepackage{inconsolata}

\usepackage{graphicx}

\title{Confidence-guided Refinement Reasoning for \\ Zero-shot Question Answering}

\author{Youwon Jang$^{1}$ ~~~~~ Woo Suk Choi$^{1}$ ~~~~~ Minjoon Jung$^{1}$ \\
{\bf Minsu Lee$^{2*}$ ~~~~~ Byoung-Tak Zhang$^{1}$}\thanks{Corresponding author}\\
$^1$Seoul National University ~~~~~ $^2$Sungshin Women’s University\\
\small{\texttt{\{ywjang, wschoi, mjjung\}@bi.snu.ac.kr}}, ~
\small{\texttt{mslee@sungshin.ac.kr}}, ~
\small{\texttt{btzhang@bi.snu.ac.kr}}
}

\usepackage{graphicx}
\usepackage{amsmath}
\usepackage{amssymb}
\usepackage{booktabs}

\usepackage{kotex}
\usepackage{xcolor}
\usepackage{multirow}
\usepackage{booktabs}
\usepackage{mathtools} 
\usepackage{algorithm}
\usepackage{algorithmic}
\usepackage{colortbl}
\usepackage{listings}
\usepackage{bbding}
\usepackage{hyperref}

\newcommand{\blue}[1]{{\color{blue}#1}}
\newcommand{\grey}[1]{\cellcolor[HTML]{eeeeee}{#1}}

\newcommand{\Framework}{{Confidence-guided Refinement Reasoning}}
\newcommand{\abbr}{{C2R}}
\newcommand{\singlesubqa}{\textsc{SingleSubQA}}
\newcommand{\everysubqa}{\textsc{EverySubQA}}
\newcommand{\subqajudge}{\textsc{LLMVerified}} 

\newcommand{\bank}{{sub-QA bank}}
\newcommand{\VQA}{{VideoQA}}
\newcommand{\IQA}{{ImageQA}}

\usepackage[utf8]{inputenc}
\usepackage[T1]{fontenc}
\usepackage{amsmath, amssymb}
\usepackage{xspace}
\usepackage{graphicx}
\usepackage{makecell}
\usepackage{multirow}
\usepackage[table]{xcolor}

\definecolor{NavyBlue}{RGB}{0,0,128}

\hypersetup{
  colorlinks=true,
  linkcolor=black,
  citecolor=black,
  urlcolor=black
}

\begin{document}
\maketitle
\begin{abstract}

We propose Confidence-guided Refinement Reasoning (C2R), a novel training-free framework applicable to question-answering (QA) tasks across text, image, and video domains. C2R strategically constructs and refines sub-questions and their answers (sub-QAs), deriving a better confidence score for the target answer. C2R first curates a subset of sub-QAs to explore diverse reasoning paths, then compares the confidence scores of the resulting answer candidates to select the most reliable final answer. Since C2R relies solely on confidence scores derived from the model itself, it can be seamlessly integrated with various existing QA models, demonstrating consistent performance improvements across diverse models and benchmarks. Furthermore, we provide essential yet underexplored insights into how leveraging sub-QAs affects model behavior, specifically analyzing the impact of both the quantity and quality of sub-QAs on achieving robust and reliable reasoning.

\end{abstract}


\section{Introduction}
\label{sec:intro}

\begin{figure*}[t]
    \centering
    \includegraphics[width=\textwidth]{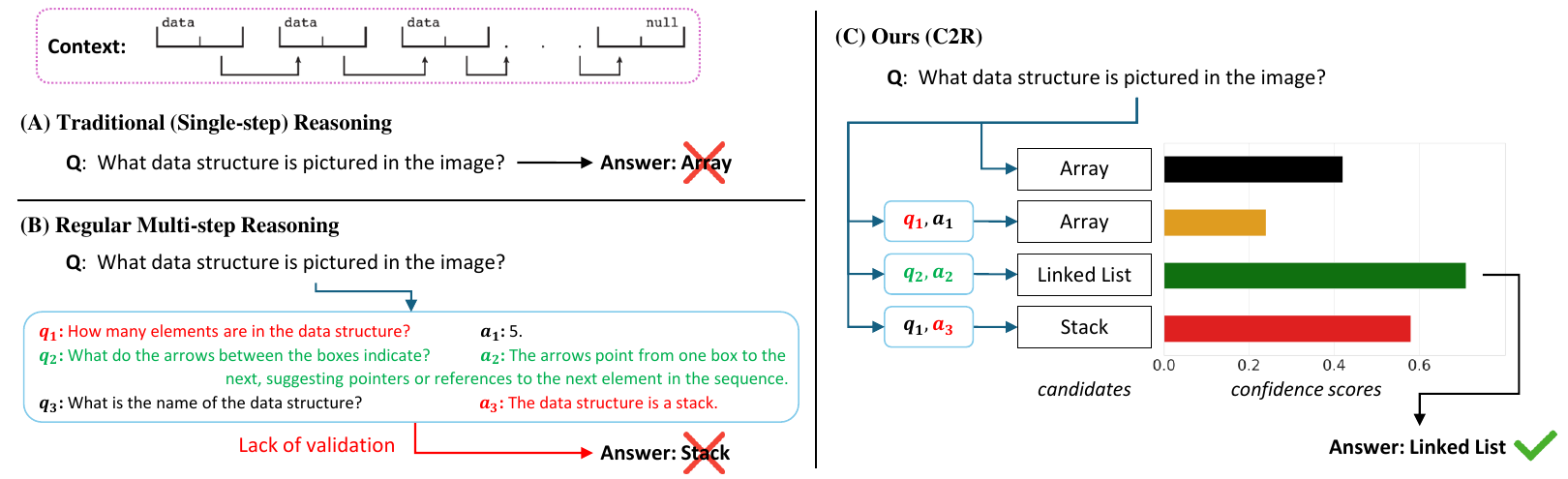}
    \caption{\textbf{Comparison of existing QA methods.} \textbf{(A):} Traditional single-step reasoning. \textbf{(B):} Multi-step reasoning via question decomposition, typically employed for complex problems, but often overlooking inaccuracies in generated sub-QAs (highlighted in red text). \textbf{(C):} \abbr~minimizes the risk of relying on \textit{unverified} sub-QAs by selecting the most confident answer among multiple candidates based on their confidence scores.
  }
    \label{fig:motivation}
\end{figure*}

Question-answering (QA) is a fundamental reasoning task that demands a deep understanding of given content and has been 
explored across various domains, including texts~\cite{hendrycks2020measuring, wang2024mmlu}, images~\cite{malinowski2014multi, zhu2016visual7w}, and videos~\cite{xiao2021next, engin2023zero}.
Traditional approaches treat QA tasks as a single-step reasoning problem, directly generating answers without intermediate analysis (Figure~\ref{fig:motivation}A). However, answers to complex questions may not be directly inferable and 
often necessitate multi-step reasoning.
As illustrated in Figure~\ref{fig:motivation} (B), multi-step reasoning methods~\cite{Uehara_2022_CVPR, wang2022understanding, khan2023exploring, yao2023tree, besta2024graph} have been proposed that
decompose the main question into several sub-question-answer pairs (sub-QAs) and subsequently utilize them to derive the final answer.

However, we argue that sub-QAs do not always enhance QA reasoning; their indiscriminate use can adversely affect and distract the model's inference procedure. Our investigation reveals that when sub-QAs are irrelevant to the main question or paired with inaccurate answers, they introduce noise into the reasoning process, ultimately degrading answer quality. Despite these risks, prior works~\cite{Uehara_2022_CVPR, khan2023exploring, you2023idealgptiterativelydecomposingvision, liao2024align} often incorporate sub-QAs without sufficient verification or refinement, as their relevance to the main question is neither assessed nor guaranteed.


To this end, we propose \Framework~(\abbr), a framework comprising three core components: Generator, Refiner, and Answer Selector. The Generator decomposes a main question into multiple sub-questions and generates corresponding sub-answers. The Refiner then selectively curates a subset of these sub-QAs to explore diverse reasoning paths
Each reasoning path yields an answer candidate accompanied by a confidence score, obtained directly from the model. Finally, the Answer Selector determines the most appropriate answer by comparing these confidence scores, including the score of the single-step answer that does not rely on any sub-QAs.


%
As C2R only relies on self-assessed confidence scores, it is inherently training-free and readily adaptable to various QA models.
By extensive experiments across various QA tasks, C2R demonstrates consistent improvements on a range of state-of-the-art models~\cite{yang2024qwen2, team2025gemma, li2024llavaonevisioneasyvisualtask, hurst2024gpt}. To provide further insights, we examine the impact of visual information on C2R performance and determine the optimal number of sub-QAs for achieving robust performance. We also reveal the \textit{confidence inflation} problem, where utilizing sub-QAs increases the confidence scores for answer candidates even when they are incorrect, underscoring the need for careful usage of sub-QAs. To mitigate this, we set confidence thresholds to reliably select the final answer and analyze their impact on model behavior and performance.

To summarize, our contributions are threefold:
\begin{itemize}
    \item We propose the \Framework~(\abbr) framework, which strategically utilizes sub-QAs and guides the model in selecting final answers based on self-assessed confidence scores. 
    \item We demonstrate that \abbr~consistently achieves improvements across five models and five benchmarks in zero-shot settings.
    \item We provide insights into often-overlooked issues concerning the impact of utilizing irrelevant or incorrect sub-QAs within multi-step reasoning QA methodologies.
\end{itemize}

\section{Related works}
\label{sec:related}

\begin{figure*}[t]
    \centering
    \includegraphics[width=\textwidth]{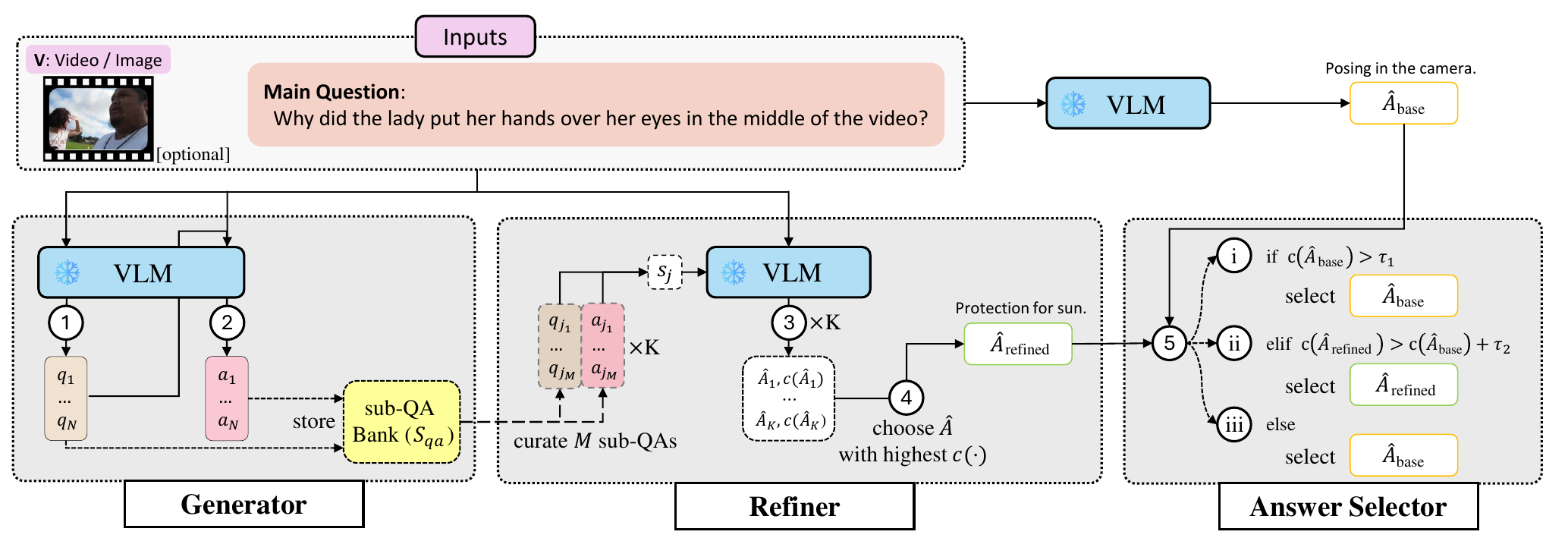}
    \caption{
    \textbf{An overview of \Framework~(C2R)}. Given content $V$ and main question $Q$,
    Generator constructs $N$ sub-QAs.
    Refiner then curates $K$ subsets, each consisting of $M$ sub-QAs, to generate answer candidates with their corresponding confidence scores.
    Finally, Answer Selector chooses one as the final answer from $\hat{A}_{\text{base}}$ and $\hat{A}_{\text{refined}}$ using the confidence thresholds $\tau_1$ and $\tau_2$. Vision-Language Model (VLM) is freezed.
    }
    \label{fig:architecture}
\end{figure*}

\subsection{Question answering} 
\label{sec:related:QA}

Question-answering (QA) tasks span various modalities, including text, image, and video.
Text-only QA evaluates a model's reasoning ability and factual knowledge across diverse domains. 
A range of benchmarks~\cite{hendrycks2020measuring, wang2024mmlu, geva2021didaristotleuselaptop} evaluate a model's factual knowledge, complex reasoning capabilities, and ability to perform multi-hop inference.
\IQA~\cite{malinowski2014multi, antol2015vqa}, also known as Visual Question Answering, involves providing an accurate natural language answer to a question based on a given image. 
Recently, to assess more profound knowledge and reasoning ability, MMMU~\cite{yue2024mmmumassivemultidisciplinemultimodal} includes single- and multi-image questions in both multiple-choice and open-ended formats across diverse academic domains.
\VQA~\cite{tapaswi2016movieqa, xiao2021next} extends the complexity by incorporating temporal dynamics. Datasets like EgoSchema~\cite{mangalam2023egoschema} present long-form videoQA challenges, requiring models to comprehend and reason over extended video sequences. 
To address QA tasks across these diverse domains, powerful foundation models~\cite{maaz2023video, li2024llavaonevisioneasyvisualtask, yang2024qwen2, team2025gemma} have been proposed, leveraging techniques like large-scale pretraining.
In this work, we propose a general approach to improving QA performance across diverse domains by building upon unified models capable of handling text and vision modalities.

\subsection{Multi-step reasoning for QA}
\label{sec:related:multi_step_reasoning}
To tackle questions requiring complex reasoning, many multi-step approaches have recently been proposed. Modular approaches~\cite{gupta2023visual, suris2023vipergpt, choudhury2023zero} decompose complex instructions into a sequence of sub-programs, executing them through sub-modules. Chain-of-Thought~\cite{wei2022chain} and CoT-SC~\cite{wang2022self} are prompting techniques that encourage models to generate intermediate reasoning steps before producing a final answer, significantly improving the capabilities of large language models (LLMs). However, these methods lack an automated mechanism for error correction and are thus vulnerable to cascading failures caused by errors in early steps.

To overcome this limitation, methods like ToT~\cite{yao2023tree}, GoT~\cite{besta2024graph}, and DeAR~\cite{xue2024decompose} have been proposed, where intermediate steps are organized in tree or graph structures to explore optimal reasoning paths. They also include a verification step where the sub-answers are checked by prompting the LLM directly. However, verifying each intermediate step using an LLM can be suboptimal~\cite{wang2023large} and requires additional computation.

In this paper, we propose a confidence-based approach that minimizes the impact of inaccurate sub-QAs without directly evaluating each one, thereby improving both efficiency and robustness.
We also analyze the effectiveness of directly assessing sub-QAs with an LLM in Appendix~\ref{sec:appendix:additional_analyses:subqa_judge}.

\section{\abbr~Framework}
\label{sec:methods}

In this section, we begin by explaining how the model derives confidence scores from its answers (\S\ref{sec:methods:preliminaries}) and then outline the basic inference process in QA tasks (\S\ref{sec:methods:base}). Subsequently, we introduce the three core components of \abbr: Generator, Refiner, and Answer Selector (\S\ref{sec:methods:subqa_generator}--\S\ref{sec:methods:answer_selector}).
Figure~\ref{fig:architecture} illustrate our \abbr~framework. 
In the following explanation, $V$ can be an image or a video.

\subsection{Confidence of the answer}
\label{sec:methods:preliminaries}

When the model $f$ generates an answer $\hat{A}$ given context $V$ and a main question $Q$, we measure its confidence as $c(\hat{A})$, where $c(\cdot)$ is the confidence scoring function.
First, for the answer sequence $\hat{A}=\{y_1, \dots, y_L\}$, we apply the softmax function to the logits and denote the probability of the selected token $y_i$ as $p_i$. This yields a sequence of probabilities $[p_1, \dots, p_L]$ corresponding to the generated tokens.
Then, we compute the confidence score as $c(\hat{A}) := \text{min}^L_{i=1} \ p_i$.
We adopt the minimum probability across the answer tokens as a conservative strategy~\cite{geng2023survey, kim2024videoicl}, under the assumption that the least confident token may indicate a potential failure point. 
Additional confidence metrics are discussed in Appendix~\ref{sec:appendix:additional_analyses:calibration}.

\subsection{Base inference process} 
\label{sec:methods:base}
Given context $V$ and a main question $Q$, the corresponding answer $\hat{A}_\text{base}$ in an open-ended QA task is obtained from the model $f$ as:
\begin{equation}
   \hat{A}_\text{base} = f(V, Q).
  \label{eq:base_OE}
\end{equation}
For a multiple-choice QA task, the model $f$ selects the answer from a set of answer options $\{A_\text{option}\}$ as: 
\begin{equation}
   \hat{A}_\text{base} = f(V, Q, \{ A_\text{option} \} ).
  \label{eq:base_MC}
\end{equation}
We refer to $\hat{A}_\text{base}$ as the \textit{base answer}. If the confidence of the base answer (i.e., $c(\hat{A}_{\text{base}})$) is sufficiently high, the subsequent steps are skipped. We explain this in \S\ref{sec:methods:answer_selector}.

\begin{algorithm}[tb]
\caption{Confidence-guided Refinement Reasoning}
\label{alg:algorithm}
\textbf{Input}: $V$ (optional): video or image(s), $Q$: main question (related to $V$) \\ 
\textbf{Parameter}: $N, M, K, \tau_1, \tau_2$\\
\textbf{Output}: $\hat{A}$: answer to main question $Q$\\
\begin{algorithmic}[1] 
\STATE $\hat{A}_\text{base}, c(\hat{A}_\text{base}) = f(V, Q)$ 
\IF {$c(\hat{A}_\text{base}) \geq \tau_1$}
    \STATE \textbf{return} $\hat{A}_\text{base}$
\ENDIF
    
\STATE $S_{qa} \leftarrow \{\}$

\FOR{$i \gets 1$ to $N$}
    \STATE $q_i = f(V, Q), ~a_i = f(V, q_i)$ 
    \STATE $S_{qa} = S_{qa} \cup \{(q_i, a_i)\}$ 
\ENDFOR

\FOR{$j \gets 1$ to $K$}
    \STATE curate $s_j:=(q_1^j, a_1^j), ..., (q_M^j, a_M^j)$ \\ from $S_{qa}$ 
    \STATE $\hat{A}_j, c(\hat{A}_j) = f(V, Q | s_j)$ 
\ENDFOR    

\STATE $\text{idx} = \underset{j}{\mathrm{argmax}} \ c(\hat{A}_j)$
\STATE $\hat{A}_\text{refined} = \hat{A}_\text{idx}$

\IF{$c(\hat{A}_\text{refined}) \ge c(\hat{A}_\text{base}) + \tau_2$}
    \STATE \textbf{return} $\hat{A}_\text{refined}$
\ELSE
    \STATE \textbf{return} $\hat{A}_\text{base}$
\ENDIF
    
\end{algorithmic}
\end{algorithm}

\subsection{Generator}
\label{sec:methods:subqa_generator}
Given $V$ and $Q$, the Generator constructs sub-questions and their corresponding answers.
Formally, the model $f$ generates $N$ sub-QAs as:
\begin{equation}
  q_i \sim f(V, Q),~a_i =  f(V, q_i), \quad 1 \leq i \leq N.
  \label{eq:subQA}
\end{equation} 
We refer to this set of generated sub-QAs as the \bank, denoted $S_{qa}$: 
\begin{equation}
    S_{qa} = \{ (q_1, a_1), (q_2, a_2), \dots, (q_N, a_N) \}.
    \label{eq:subQA_bank}
\end{equation}

\subsection{Refiner}
\label{sec:methods:refiner}

We introduce the Refiner to avoid irrelevant sub-QAs leading the model to an incorrect conclusion. 
First, the Refiner curates $K$ sub-QA subsets from the \bank~$S_{qa}$ while satisfying the following conditions:
1) The indices of sub-QA pairs within each subset should be unique.
2) Any two curated subsets may share some sub-QA pairs, but they must not be identical in composition.

Formally, the $j$-th curated sub-QA subset is denoted as $\{(q_1^j, a_1^j), \dots, (q_M^j, a_M^j)\}$, where $1 \leq j \leq K$. We then concatenate the selected sub-QAs to form the curated context $s_j$:
\begin{equation}
  s_j = [ (q_1^j, a_1^j) \ || \ \dots \ || \ (q_M^j, a_M^j) ],
  \label{eq:curate}
\end{equation}
where
$\forall i, k \in \{1, \dots, M\}, \; (q_i^j, a_i^j) \neq (q_k^j, a_k^j)$ for $i \neq k$, and $||$ denotes the concatenation operator. We then feed the $s_j$ as additional input to the model to generate a candidate answer $\hat{A}_j$:
\begin{equation}
  \hat{A}_j = f(V, Q \ | \ s_j)
  \label{eq:gen_refined}
\end{equation}

As a result, we obtain $K$ answer candidates $\{\hat{A}_j\}$, where $1 \leq j \leq K$.\footnote{We do not necessarily explore all $K$ reasoning paths for every question, as early stopping is possible when sufficient confidence is achieved. This is further explained in Appendix~\ref{sec:appendix:additional_analyses:cost}.}
Finally, the refined answer, $\hat{A}_{\text{refined}}$, is selected as the candidate from $\{\hat{A}_j\}_{j=1}^K$ that has the highest confidence score.


\subsection{Answer Selector}
\label{sec:methods:answer_selector}

Although $\hat{A}_{\text{refined}}$ is obtained using sub-QAs that are likely relevant to $V$ and $Q$, it remains uncertain whether it is the correct answer. To improve the reliability of the inferred answer, we design the Answer Selector to guide the model toward more accurate conclusions. Our Answer Selector follows two principles:

\noindent\textbf{Principle 1.} For $V$ and $Q$ that are relatively easy to understand and answer, overcomplicating the task can distract the model and hinder it from identifying the correct answer~\cite{shi2023large}. Therefore, if the base answer $\hat{A}_{\text{base}}$ exhibits a sufficiently high confidence score (i.e., $c(\hat{A}_{\text{base}}) \geq \tau_1$), we choose it as the final answer, and subsequent refinement steps are skipped. We use a confidence threshold $\tau_1$ for this purpose.

\noindent\textbf{Principle 2.} If $c(\hat{A}_{\text{base}}) < \tau_1$, the Answer Selector considers both $\hat{A}_{\text{base}}$ and $\hat{A}_{\text{refined}}$. However, a naive comparison of $c(\hat{A}_{\text{base}})$ and $c(\hat{A}_{\text{refined}})$ yields only marginal improvement. We empirically observe that using sub-QAs tends to inflate confidence scores compared to $c(\hat{A}_{\text{base}})$, even when incorrect sub-QAs are provided, potentially leading to erroneous selections. To address this, we introduce an additional confidence threshold $\tau_2$. Specifically, $\hat{A}_{\text{refined}}$ is selected if its confidence score $c(\hat{A}_{\text{refined}})$ satisfies $c(\hat{A}_{\text{refined}}) \geq c(\hat{A}_{\text{base}}) + \tau_2$; otherwise, $\hat{A}_{\text{base}}$ is chosen.

These principles guide the selection of the more reliable answer between $\hat{A}_{\text{base}}$ and $\hat{A}_{\text{refined}}$. The full procedure is listed in Algorithm~\ref{alg:algorithm}, and we analyze the effect of confidence thresholds in \S\ref{sec:analysis:confidence}.

\begin{table*}[!ht]
    \newlength{\anotherShiftdown} 
    \setlength{\anotherShiftdown}{-0.25em}
    \centering
    \resizebox{1.0\linewidth}{!}{
    \begin{tabular}{l|l|@{\hskip 6pt}lccccccccccccccc@{\hskip 6pt}}
        \toprule
        \multirow{2}{*}{\textbf{Backbone}} & \multirow{2}{*}{\textbf{Size}}& \multirow{2}{*}{\textbf{Reasoning Method}} &
        \multirow{2}{*}{\boldmath{$N$}} & \multirow{2}{*}{\boldmath{$M$}} & \multirow{2}{*}{\boldmath{$K$}} & 
        \multicolumn{3}{c}{\textbf{Text-only QA}} & \multicolumn{1}{c}{\textbf{ImageQA}} & \multicolumn{1}{c}{\textbf{VideoQA}}  \\
        \cmidrule(lr){7-9} \cmidrule(lr){10-10} \cmidrule(lr){11-11}
        & & & & & & \textbf{MMLU} & \textbf{MMLU-Pro} & \textbf{StrategyQA} & \textbf{MMMU} & \textbf{EgoSchema}  \\ \midrule

        \multirow{6}{*}{Qwen2.5} & \multirow{6}{*}{7B} & 
        Baseline                       & - & - & - & 67.1          & 38.6          & 62.6          & \underline{50.2}    & 68.0          \\
        && \textsc{SingleSubQA}       & 1 & 1 & 1 & \textcolor{blue}{67.5}          & \textcolor{blue}{42.9}          & 62.6          & \textcolor{red}{47.6}           & \textcolor{blue}{\underline{69.2}} \\
        && \textsc{EverySubQA}        & 5 & 5 & 1 & \textcolor{blue}{\underline{67.7}} & \textcolor{blue}{42.6}          & \textcolor{blue}{\underline{65.3}} & \textcolor{red}{47.6}           & \textcolor{blue}{69.0}          \\
        && \subqajudge    (sub-A)   & 5 & 2 & 1 & \textcolor{blue}{67.3}          & \textcolor{blue}{\underline{43.2}} & \textcolor{blue}{64.5}          & \textcolor{red}{49.4}           & \textcolor{red}{67.0}           \\
        && {\textbf{\abbr (Ours)}}  & 5 & 2 & 4 & \textcolor{blue}{\textbf{69.2}} & \textcolor{blue}{\textbf{44.3}} & \textcolor{blue}{\textbf{65.7}} & \textcolor{blue}{\textbf{51.9}} & \textcolor{blue}{\textbf{71.5}} \\
        && \multicolumn{1}{l}{\blue{\grey{~$\Delta$}}} & \grey{} & \grey{} & \grey{} & \blue{\bf \grey{+2.1}} & \blue{\bf \grey{+5.7}} & \blue{\bf \grey{+3.1}}& \blue{\bf \grey{+1.7}} & \blue{\bf \grey{+3.5}} \\
        \midrule

        \multirow{6}{*}{LLaVA-Onevision} & \multirow{6}{*}{7B} & 
        Baseline                       & - & - & - & \underline{66.2}    & 35.7          & 64.5          & 45.1          & 43.0          \\
        && \textsc{SingleSubQA}       & 1 & 1 & 1 & \textcolor{red}{64.3}           & \textcolor{blue}{36.6}          & \textcolor{blue}{67.8}          & \textcolor{red}{44.1}           & \textcolor{red}{40.2}           \\
        && \textsc{EverySubQA}        & 5 & 5 & 1 & \textcolor{red}{63.4}           & \textcolor{blue}{\underline{37.0}} & \textcolor{blue}{67.6}          & \textcolor{blue}{46.1}          & \textcolor{red}{39.8}           \\
        && \subqajudge    (sub-A)   & 5 & 2 & 1 & \textcolor{red}{65.1}           & \textcolor{blue}{36.9}          & \textcolor{blue}{\underline{68.2}} & \textcolor{blue}{\underline{47.2}} & \textcolor{blue}{\underline{45.2}} \\
        && {\textbf{\abbr (Ours)}}  & 5 & 2 & 4 & \textcolor{blue}{\textbf{67.3}} & \textcolor{blue}{\textbf{38.4}} & \textcolor{blue}{\textbf{68.3}} & \textcolor{blue}{\textbf{47.4}} & \textcolor{blue}{\textbf{46.0}} \\
        && \multicolumn{1}{l}{\blue{\grey{~$\Delta$}}} & \grey{} & \grey{} & \grey{} & \blue{\bf \grey{+1.1}} & \blue{\bf \grey{+2.7}} & \blue{\bf \grey{+3.8}}& \blue{\bf \grey{+2.3}} & \blue{\bf \grey{+3.0}}  \\
        \midrule

        \multirow{3}{*}{Gemma 3} & \multirow{3}{*}{4B} & 
        Baseline                                   & - & - & - & 58.9          & 27.3          & 45.9          & 40.1          & 50.5          \\
        && {\textbf{\abbr (Ours)}}                 & 5 & 2 & 4 & \textcolor{blue}{\textbf{60.0}} & \textcolor{blue}{\textbf{30.5}} & \textcolor{blue}{\textbf{50.1}} & \textcolor{blue}{\textbf{42.1}} & \textcolor{blue}{\textbf{53.2}} \\
        && \multicolumn{1}{l}{\blue{\grey{~$\Delta$}}} & \grey{} & \grey{} & \grey{} & \blue{\bf \grey{+1.1}} & \blue{\bf \grey{+3.2}} & \blue{\bf \grey{+4.2}}& \blue{\bf \grey{+2.0}} & \blue{\bf \grey{+2.7}} \\
        \midrule
        
        \multirow{3}{*}{Qwen2} & \multirow{3}{*}{2B} & 
        Baseline                                   & - & - & - & 50.0          & 23.6          & \textbf{55.0} & 41.4          & 56.8          \\
        && {\textbf{\abbr (Ours)}}                 & 5 & 2 & 4 & \textcolor{blue}{\textbf{50.8}} & \textcolor{blue}{\textbf{24.7}} & \textbf{55.0} & \textcolor{blue}{\textbf{42.4}} & \textcolor{blue}{\textbf{61.0}} \\
        && \multicolumn{1}{l}{\blue{\grey{~$\Delta$}}} & \grey{} & \grey{} & \grey{} & \blue{\bf \grey{+0.8}} & \blue{\bf \grey{+1.1}} & \blue{\grey{+0.0}}& \blue{\bf \grey{+1.0}} & \blue{\bf \grey{+4.2}} \\
        \midrule
        
        \multirow{3}{*}{GPT-4o} & \multirow{3}{*}{-} & 
        Baseline                                   & - & - & - & 85.0          & -             & -             & 56.1          & 75.0          \\
        && {\textbf{\abbr (Ours)}}                 & 5 & 2 & 4 & \textcolor{blue}{\textbf{86.2}} & -             & -             & \textcolor{blue}{\textbf{58.3}} & \textcolor{blue}{\textbf{78.2}} \\
        && \multicolumn{1}{l}{\blue{\grey{~$\Delta$}}} & \grey{} & \grey{} & \grey{} & \blue{\bf \grey{+1.2}} & \blue{\grey{-}} & \blue{\grey{-}}& \blue{\bf \grey{+2.2}} & \blue{\bf \grey{+3.2}} \\

        \bottomrule
    \end{tabular}
}
    \caption{
    \textbf{Zero-shot evaluation results of \abbr.}
    We evaluate baseline models and our framework across various domains and benchmarks for comparison. $N$, $M$, and $K$ denote the number of generated sub-QAs, the number of curated sub-QAs per reasoning path, and the total number of reasoning paths, respectively.
    The difference (\blue{$\Delta$}) denotes the score improvement of \abbr~over the vanilla models. 
    \textbf{Bold text} indicates the highest score and \underline{underlined text} the second-highest. Colors show performance relative to baseline: \textcolor{blue}{blue} (higher), \textcolor{red}{red} (lower).
    }
    \label{table_main_result}
\end{table*}

\section{Experiments}
\label{sec:experiment}

In this section, we provide experimental results to demonstrate the effectiveness of our \abbr~framework. In \S\ref{sec:experiment:setup}, we outline the experimental setup, including datasets and baselines. Subsequently, we compare our framework with various multi-step reasoning methods across various QA models (\S\ref{sec:experiment:main_result}), and visualize qualitative results (\S\ref{sec:experiment:qualitative}). The prompt designs used for sub-QA generation and model inference can be found in Appendix~\ref{sec:appendix:prompt}.

\begin{figure*}[t]
    \centering
    \includegraphics[width=1.0\textwidth]{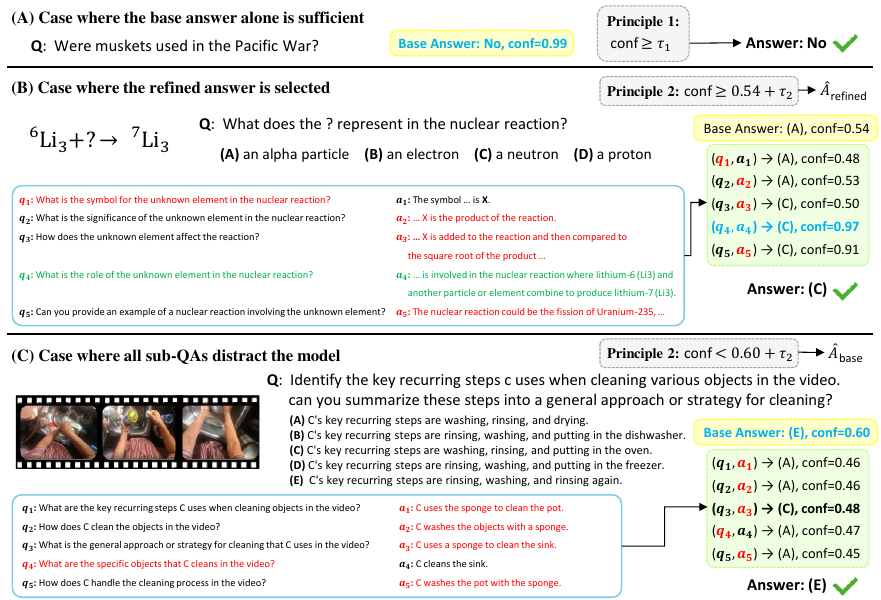}
    \caption{
    \textbf{Qualitative Results.} For simplicity, we visualize the case where $M=1$. 
    \textbf{(A)} shows a case where the base answer has a sufficiently high confidence, leading to its selection without generating refined answers.
    In \textbf{(B)}, the fourth sub-QA provides helpful information, enabling Refiner to arrive at the correct answer. With high confidence, Answer Selector then chooses the refined answer.
    In \textbf{(C)}, the refined answer is incorrect due to distraction from sub-QAs. Its low confidence leads to rejection by Answer Selector, which chooses the correct base answer.
    } 
    \label{fig:qual_result}
\end{figure*}

\subsection{Experimental setup}
\label{sec:experiment:setup}

For all experiments, we fix the number of generated sub-QAs to $N=5$ and consistently use greedy search for decoding.
All experiments are conducted on a single A6000 GPU, except for closed-source models. For closed-source models, we evaluate one benchmark per domain due to cost constraints.
Further details are provided in Appendix~\ref{sec:appendix:settings}.

\noindent\textbf{Datasets.}
We evaluate our framework on five diverse benchmarks for Text-only, Image, and Video QA, covering various domains for a comprehensive evaluation.
For Text-only QA, we use the following three datasets: MMLU~\cite{hendrycks2020measuring}, MMLU-Pro~\cite{wang2024mmlu}, and StrategyQA~\cite{geva2021didaristotleuselaptop}, to cover different reasoning demands.
MMLU consists of various academic and professional tasks requiring broad world knowledge, and MMLU-Pro extends the number of answer candidates from four to ten, making the model's decision more challenging. StrategyQA assesses a model’s ability to perform implicit, multi-hop reasoning by requiring strategic inference over unstated facts.
For \IQA, we use MMMU~\cite{yue2024mmmumassivemultidisciplinemultimodal}, which evaluates deliberate reasoning across diverse image types with college-grade questions spanning six disciplines. Its breadth and depth make it a strong benchmark for assessing domain-specific visual understanding.
For \VQA, we employ EgoSchema~\cite{mangalam2023egoschema}, which focuses on long-form egocentric videos depicting naturalistic human activities. This dataset assesses comprehension of extended temporal contexts and complex behaviors, offering a rigorous testbed for evaluating advanced video reasoning capabilities.


\noindent\textbf{Backbones.}
We apply our framework to several vision-language models capable of performing QA tasks: Qwen2.5-VL-7B~\cite{yang2024qwen2}, Gemma-3-4B~\cite{team2025gemma}, and LLaVA-Onevision-7B~\cite{li2024llavaonevisioneasyvisualtask}.
To broaden the evaluation spectrum, we also include a relatively small open-source model, Qwen2-VL-2B~\cite{wang2024qwen2vlenhancingvisionlanguagemodels}, and a state-of-the-art very large closed-source model, GPT-4o~\cite{hurst2024gpt}. 
Please refer to Appendix~\ref{sec:appendix:model_details} for further details.

\noindent\textbf{Reasoning Methods.} To further analyze the effectiveness of our framework, we introduce three additional reasoning methods. \singlesubqa~is a baseline in which the answer is derived using only a single sub-QA, which is always accepted without further selection~\cite{khan2023exploring}. \everysubqa~uses all $N$ generated sub-QAs to answer the main question. \subqajudge~introduces a sub-QA verification step instead of exploring multiple reasoning paths (i.e., $K=1$). It selects the $M$ most helpful sub-QAs and generates a single refined answer using only these selected sub-QAs.

\subsection{Zero-shot QA results}
\label{sec:experiment:main_result}



Although \singlesubqa~and \everysubqa~often outperform the baseline, their improvements are unstable, leading to a substantial performance drop. For instance, \singlesubqa~improves LLaVA-Onevision on Text-only QA datasets, but decreases its performance on both MMMU and EgoSchema. Furthermore, despite leveraging a larger number of sub-QAs, \everysubqa~frequently fails to yield any improvement and even underperforms than \singlesubqa~on EgoSchema across different backbones.
These results highlight that naively incorporating sub-QAs can negatively impact the model's performance. Moreover, simply increasing the number of intermediate reasoning steps is ineffective without proper verification.

We also compare \abbr~with \subqajudge~that sub-answers are evaluated by the underlying LLM (or VLM). 
\subqajudge~can show performance gains in some cases (e.g., LLaVA-Onevision's improvements on MMMU and EgoSchema), they can also lead to degraded performance on other benchmarks (e.g., MMLU). This indicates that such evaluation strategies may struggle to reliably identify truly helpful sub-QAs. In contrast, our approach increases the chances of accurate reasoning by aggregating signals from multiple reasoning paths—even when some paths include less reliable sub-answers. As a result, our method yields more consistent and robust performance improvements across diverse tasks.
Related experiments on identifying the most helpful sub-questions or evaluating the overall quality of the sub-QAs set are provided in Appendix~\ref{sec:appendix:additional_analyses:subqa_judge}.



Overall, our framework consistently improves performance over the baselines across different backbones and tasks, demonstrating its effectiveness. Additionally, we confirm that C2R performs well even with smaller backbone models: Gemma 3 and Qwen2, as well as with the closed-sourced model: GPT-4o, demonstrating its flexibility.

\subsection{Qualitative results}
\label{sec:experiment:qualitative}

Figure~\ref{fig:qual_result} illustrates the behavior of our \abbr~framework, across three distinct scenarios, showcasing its adaptability and robustness.
In case (A), the base answer achieves a sufficiently high confidence score to be selected without invoking the Refiner. This demonstrates that \abbr~can efficiently recognize when refinement is unnecessary, thereby preserving computational resources while maintaining accuracy.
Meanwhile, case (B) exemplifies a situation where the base answer's confidence is initially insufficient. Here, the fourth sub-QA helps the Refiner find the correct answer, which the Answer Selector chooses based on high confidence. This highlights \abbr's ability to extract valuable signals from sub-QAs even when some are uninformative.
However, there might be cases like (C) where the generated sub-QAs do not effectively support refinement. In this instance, the model falls back to the base answer, as it retains a higher confidence score than the refined options. This case illustrates the safety mechanism of the \abbr, where unreliable refinement paths are effectively discarded. 
Overall, these cases highlight the adaptability of C2R, which can selectively refine when needed and fall back to the base answer when appropriate, ensuring both robustness and efficiency.
For a detailed analysis of representative failure cases, please refer to Appendix~\ref{sec:appendix:failure_analysis}.

\section{Analysis}
\label{sec:analysis}

In this section, we analyze our method on various axes. 
First, we investigate the optimal number of sub-QAs ($M$) used to generate each answer candidate. We then analyze the effectiveness of our framework on multimodal tasks by conducting a blind test that excludes access to visual information. Finally, we provide an analysis of the confidence score and its threshold. We also provide other aspects, such as computational cost, in Appendix~\ref{sec:appendix:additional_analyses}.

\subsection{The number of curated sub-QAs}
\label{sec:analysis:M}

\begin{table}[t]
\small{
\centering
    \begin{tabular}{c ccc}
    \toprule
    \boldmath{$M$} & \textbf{MMLU} & \textbf{MMMU} & \textbf{EgoSchema} \\  
    \midrule
    -        & 67.1 & 50.2 & 68.0 \\
    \midrule
    1        & +1.3 & +0.9 & +1.0 \\
    2        & +2.1 & \textbf{+1.7} & \textbf{+3.5} \\
    3        & \textbf{+2.2} & +1.1 & +3.0 \\
    4        & +2.1 & +1.4 & +1.5 \\
    \bottomrule
    \end{tabular}
    \caption{Performance gain with varying numbers of curated sub-QAs ($M$) per answer candidate.
    Results are based on Qwen2.5-VL with $N=5$.}
    \label{table_ablation_M}
    }
\end{table}



\begin{table}[t]
\newcommand{\ft}[1]{\textcolor{gray}{#1}}
\centering
\resizebox{\linewidth}{!}{
    \begin{tabular}{ll|cc|cc}
    \toprule
    \multirow{2}{*}{\textbf{Model}} & & \multicolumn{2}{c}{\textbf{MMMU}} & \multicolumn{2}{c}{\textbf{EgoSchema}} \\
    && \textbf{Blind} & \textbf{Std.} & \textbf{Blind} & \textbf{Std.} \\
    \midrule
    \multirow{3}{*}{Qwen2.5-VL-7b}      & Baseline                  & 38.9 & 50.2 & 27.3 & 68.0 \\
                                        & \textbf{\abbr (Ours)}     & 41.0 & 51.9 & 30.3 & 71.5 \\
                                        & \grey{\blue{~$\Delta$}}   & \grey{\blue{\textbf{+2.1}}} & \grey{\blue{+1.7}} & \grey{\blue{+3.0}} & \grey{\blue{\textbf{+3.5}}} \\
    \midrule
    \multirow{3}{*}{LLaVA-Onevision-7b} & Baseline                  & 43.2 & 45.1 & 35.5 & 42.0 \\
                                        & \textbf{\abbr (Ours)}     & 43.4 & 47.4 & 38.0 & 45.8 \\
                                        & \grey{\blue{~$\Delta$}}   & \grey{\blue{+0.2}} & \grey{\blue{\textbf{+2.3}}} & \grey{\blue{+2.5}} & \grey{\blue{\textbf{+3.8}}} \\
    \midrule
    \multirow{3}{*}{Gemma-3-4b}         & Baseline                  & 34.0 & 40.1 & 23.3 & 50.5 \\
                                        & \textbf{\abbr (Ours)}     & 35.9 & 42.1 & 24.5 & 53.2 \\
                                        & \grey{\blue{~$\Delta$}}   & \grey{\blue{+1.9}} & \grey{\blue{\textbf{+2.0}}} & \grey{\blue{+1.2}} & \grey{\blue{\textbf{+2.7}}} \\
    \bottomrule
    \end{tabular}
}
\caption{To validate the impact of visual inputs, we evaluate under a blind setting, where QA is performed without access to visual content. `Std.' refers to the standard setting. \abbr~achieves greater improvements, confirming its effective use of visual information. 
}
\label{table_blind}
\end{table}


\begin{figure*}[t]
    \centering
    \includegraphics[width=\textwidth]{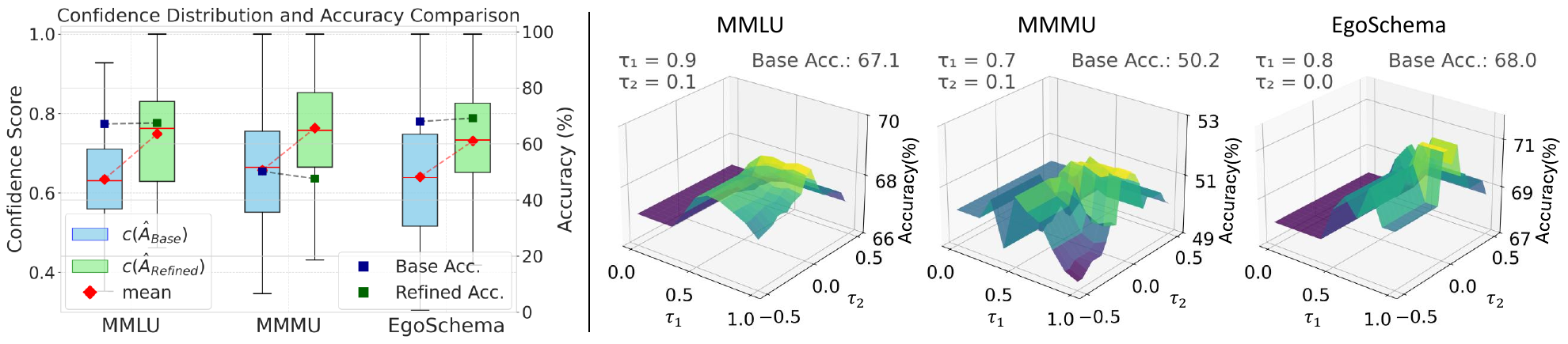}
  \caption{
    \textbf{Left:} Comparison of accuracy and average confidence scores for base and refined answers. While the accuracy difference is marginal, the average confidence score for refined answers increases significantly compared to base answers.
    \textbf{Right:} Accuracy results corresponding to different confidence threshold values for $\tau_1, \tau_2$. `Base Acc.' denotes the accuracy of the baseline model (Qwen2.5-VL).
  }
  \label{fig:conf_and_tau}
\end{figure*}

Table~\ref{table_ablation_M} shows the importance of an appropriate number of sub-QA pairs (i.e., $M$) when curating. Although all cases show improvements, curating with too small ($M=1$) or too many ($M=4)$ sub-QAs results in suboptimal performance gains. We choose $M=2$ as it demonstrates the best performance. We provide additional analysis on varying the number of refined reasoning paths (i.e., $K$) in Appendix~\ref{sec:appendix:additional_analyses:K}.

\subsection{Impact of visual input on QA reasoning}
\label{sec:analysis:visual_input}
To investigate whether C2R genuinely references visual content during utilizing sub-QAs, 
we compare its performance against the baseline under two conditions: standard and blind, depending on the presence or absence of visual inputs.
In Table~\ref{table_blind}, most cases show that the performance improvements are more pronounced in the standard settings than in the blind settings, confirming that our method effectively incorporates visual information.
We also compare C2R with different prompting methods for Text-only QA: Chain-of-Thought~\cite{wei2022chain} (CoT) and CoT-SC~\cite{wang2022self}, in Appendix~\ref{sec:appendix:additional_analyses:CoT}. 

\subsection{Confidence inflation problem} 
\label{sec:analysis:confidence}

While confidence scores are crucial for selecting the final answer, we observe that using sub-QAs often increases confidence scores even for incorrect answer candidates, leading to inaccurate reasoning.
We call this problem as \textit{confidence inflation}.
As illustrated in Figure~\ref{fig:conf_and_tau} (Left), although the accuracy gap between base and refined answers is minimal (-0.3\% on average), 
the mean confidence of refined answers is significantly inflated--by an average of 0.11.
We hypothesize that the model tends to overestimate potentially incorrect information and reasons based on it. 
Consequently, adding sub-QAs to the input shifts the distribution of confidence scores, making it unreliable to rely solely on high confidence scores in refined answers.
Due to this confidence inflation, the comparison between the base and refined answers requires careful handling, motivating the use of a confidence threshold $\tau_2$ during the answer selection process. 
For a more detailed analysis of the criteria for selecting the best answer, see Appendix~\ref{sec:appendix:additional_analyses:calibration}.

\subsection{Threshold analysis}
\label{sec:analysis:tau}
We report the results with the confidence thresholds $\tau_1$ and $\tau_2$ via a grid search on the validation sets (see Appendix~\ref{sec:appendix:settings}). To ensure these hyperparameters are not overfitted and to understand their impact, we conduct a sensitivity analysis on the test set. 
The performance landscape in Figure~\ref{fig:conf_and_tau} (Right) offers insights into benchmark characteristics. A higher optimal $\tau_1$ suggests that base answers are often insufficient, necessitating more complex reasoning. Conversely, a higher optimal $\tau_2$ indicates that refined answers from sub-QAs are less reliable than base answers for that benchmark. For instance, on MMMU, a strategy that aggressively favors refined answers (e.g., high $\tau_1$, low $\tau_2$) leads to a performance drop compared to the baseline. This highlights the importance of our selective refinement mechanism, as indiscriminately incorporating all sub-QAs can degrade overall accuracy.

Furthermore, our analysis reveals that the framework's performance is stable across a range of threshold values. We found that using a fixed threshold pair ($\tau_1=0.7, \tau_2=0.1$) across all benchmarks results in performance nearly identical to that achieved with individually optimized thresholds. As shown in Table~\ref{tab:fixed_threshold}, the performance difference is marginal, demonstrating the generalizability of our approach and reducing the need for extensive per-dataset tuning.
This analysis confirms the robustness of our chosen thresholds.

\begin{table}[t]
\centering
\resizebox{\linewidth}{!}{
    \begin{tabular}{lccc}
    \toprule
    \textbf{Model} & \textbf{MMLU} & \textbf{MMMU} & \textbf{EgoSchema} \\
    \midrule
    Baseline & 67.1 & 62.6 & 68.0 \\
    C2R (Optimal $\tau$) & 69.2 & 65.7 & 71.5 \\
    C2R (Fixed $\tau$) & 69.1 & 65.6 & 71.2 \\
    \bottomrule
    \end{tabular}
}
\caption{Performance comparison between optimally-tuned and fixed thresholds on test sets. The results show minimal degradation, highlighting the robustness of our method.}
\label{tab:fixed_threshold}
\end{table}


\section{Conclusion}
\label{sec:conclusion}
In this work, we address the challenge that sub-QAs do not always enhance QA reasoning. To tackle this, we propose \Framework~(\abbr), a model-agnostic framework that selectively curates sub-QAs into multiple subsets to construct diverse reasoning paths, rather than using all sub-QAs indiscriminately.
By selecting the most reliable answer from multiple candidates based on confidence scores, our approach achieves higher accuracy than prior methods.
We also find that sub-QAs—regardless of their actual relevance—tend to inflate the confidence of the resulting answers. To mitigate this, \abbr~employs appropriate confidence thresholds to guide the selection process, thereby reducing the risk of relying on incorrect sub-QAs.
Extensive experiments on various benchmarks demonstrate that \abbr~consistently improves performance across models without requiring additional training.
\section*{Limitations}

In \abbr, the Refiner identifies more reliable answers by exploring diverse reasoning paths compared to single-step reasoning, and the Answer Selector ultimately chooses the most accurate answer. While \abbr~provides remarkable performance, it does have some limitations.
First, although our framework effectively reduces the likelihood of utilizing suboptimal sub-QAs, it does not entirely prevent their generation.
Second, while the depth of the reasoning process can be extended beyond two steps, this work does not explore such multi-level refinement.
Third, since most benchmarks do not provide ground-truth sub-QAs, separate examples would need to be prepared for few-shot evaluation, which is beyond the scope of this paper.
In future work, we plan to explore methods 
to further reduce the likelihood of generating low-quality sub-QAs.

\section*{Acknowledgements}
This work was partly supported by the IITP (RS-2021-II212068-AIHub/10\%, RS-2021-II211343-GSAI/10\%, RS-2022-II220951-LBA/15\%, RS-2022-II220953-PICA/15\%), NRF (RS-2024-00353991-SPARC/15\%, RS-2023-00274280-HEI/10\%, RS-2024-00358416-AutoRL/5\%), KEIT (RS-2024-00423940/10\%), and Gwangju Metropolitan City (Artificial intelligence industrial convergence cluster development project/10\%) grant funded by the Korean government.



\bibliography{camera_ready}

\appendix
        \clearpage
\setcounter{page}{1}
\setcounter{section}{0}

We present the following details that are not included in the main manuscript:

\begin{itemize}
    \item \textbf{Additional Analyses}: We present additional analyses.
    \item \textbf{Benchmarks}: We provide detailed information on the benchmarks.
    \item \textbf{Details of Models}: We describe details of the models used in our experiments.
    \item \textbf{Details of Experiment Settings}: We present additional information on the experiment settings.
    \item \textbf{Prompt Designs}: We provide the prompts used for inference.
\end{itemize}

\section{Additional Analyses}
\label{sec:appendix:additional_analyses}

\subsection{Impact of the number of reasoning paths}
\label{sec:appendix:additional_analyses:K}
We experimentally determine the optimal number of reasoning paths (i.e., $K$).
Intuitively, 
having more reasoning paths leads to higher performance, and the results in Table~\ref{table_ablation_K} support this. Considering computational efficiency, we set the number of reasoning paths to $K=4$ for our main experiments.

\begin{table}[!ht]
\small
\centering
\begin{tabular}{@{}c|c|cccc@{}}
    \toprule
    & \boldmath{$K$} & \textbf{MMLU} & \textbf{MMMU} & \textbf{EgoSchema} \\  
    \midrule 
    Baseline & -        & 67.1 & 50.2 & 68.0 \\
    \midrule
    \multirow{4}{*}{\textbf{\abbr}}
             & 1        & +1.6 & +1.1 & +2.2 \\
             & 2        & +1.7 & +1.2 & +2.5 \\
             & 4        & +2.1 & \textbf{+1.7} & \textbf{+3.5} \\
             & 8        & \textbf{+2.2} & \textbf{+1.7} & +3.2 \\
    
    \bottomrule
\end{tabular}
\caption{Ablation of the number of reasoning paths. We denote the number of reasoning paths as $K$ and find that $K=4$ is best in our settings. All experiments are conducted using Qwen2.5.}
\label{table_ablation_K}
\end{table}

\subsection{Bootstrap significance test}
\label{sec:appendix:additional_analyses:significance_test}

To formally validate the robustness of our results, we conducted a bootstrap significance test to ascertain whether the performance gains of our \abbr~framework are statistically significant. We test against the null hypothesis ($H_0$) that there is no true performance difference between our framework and the baseline, using a significance level of $\alpha=0.05$.
We generate $B=10k$ bootstrap samples by resampling from the test set with replacement. The $p$-value is then calculated by comparing the performance gap on the original test set, $\delta_{\text{orig}}$, with the gaps observed on the bootstrap samples, $\delta_i$. The formula is as follows:
\begin{equation*}
p\text{-value} = \frac{1}{B} \sum_{i=1}^{B} \mathbb{I}(\delta_i \ge \delta_{\text{orig}})
\end{equation*}
where $\mathbb{I}(\cdot)$ is the indicator function, which returns 1 if the condition is true and 0 otherwise.

The resulting $p$-values for each model and benchmark are presented in Table~\ref{tab:p_values}. The results demonstrate that the p-values are well below the $\alpha=0.05$ threshold in nearly all cases, with minor exceptions for Qwen2-2b. Therefore, we can confidently reject the null hypothesis ($H_0$) and conclude that the performance improvements achieved by our \abbr~framework are statistically significant.

\begin{table}[t] 
\centering
\resizebox{\linewidth}{!}{
    \begin{tabular}{lccccc}
    \toprule
    \textbf{Model} & \textbf{MMLU} & \textbf{MMLU-Pro} & \textbf{StrategyQA} & \textbf{MMMU} & \textbf{EgoSchema} \\
    \midrule
    Qwen2.5 & 0.0002 & $<$0.0001 & 0.0037 & 0.0316 & 0.0013 \\
    Llava-Onevision & 0.0076 & $<$0.0001 & $<$0.0001 & 0.0004 & $<$0.0001 \\
    Gemma 3 & 0.0118 & 0.0006 & $<$0.0001 & 0.0039 & 0.0131 \\
    Qwen2 2b & 0.0390 & 0.0037 & --- & 0.1179 & 0.0027 \\
    GPT-4o & 0.0172 & --- & --- & 0.0314 & 0.0061 \\
    \bottomrule
    \end{tabular}
}
\caption{The p-values from the bootstrap significance test ($B=10k$) for the performance gains of \abbr~ over the baseline. Values below 0.05 indicate statistical significance. A value of `$<$0.0001' indicates that no bootstrap sample showed a performance gap as large as the original.}
\label{tab:p_values}
\end{table}

\begin{table}[t]
\centering
\resizebox{\linewidth}{!}{
    \begin{tabular}{l|l|cc}
    \toprule
    \textbf{Model} & \textbf{Method} & \textbf{MMLU} & \textbf{StrategyQA} \\
    \midrule
    \multirow{3}{*}{Qwen2.5} & Baseline & 67.1 & 62.6 \\
    & \abbr~(Ours) & \textbf{69.2} & \textbf{65.7} \\
    & 5-shot ICL & 69.5 & 66.1 \\
    \midrule
    \multirow{3}{*}{LLaVA-Onevision} & Baseline & 66.2 & 64.5 \\
    & \abbr~(Ours) & \textbf{67.3} & \textbf{68.3} \\
    & 5-shot ICL & 67.5 & 69.3 \\
    \bottomrule
    \end{tabular}
}
\caption{Comparison of our zero-shot \abbr~framework with a 5-shot in-context learning (ICL) baseline on the MMLU and StrategyQA benchmarks. Our training-free method achieves performance competitive with the few-shot approach.}
\label{tab:comparison_few_shot}
\end{table}

\subsection{Comparison with few-shot methods}
\label{sec:appendix:additional_analyses:few_shot}
To further contextualize the performance of our approach against methods that utilize labeled data, we conduct a new experiment comparing our zero-shot \abbr~framework with a competitive 5-shot in-context learning (ICL) baseline.

For the ICL setup, the models were provided with five demonstration examples sourced from a held-out split of the training data. The composition of these examples varied by dataset; for MMLU, the sub-QAs were generated by GPT-4o, whereas for StrategyQA, we utilized the ground-truth sub-QAs to construct the demonstrations.

The results of this comparison are presented in Table~\ref{tab:comparison_few_shot}. Our analysis shows that the proposed training-free \abbr~framework performs competitively against the 5-shot ICL method across both models and datasets. Notably, \abbr~achieves this level of performance without requiring any labeled examples for in-context demonstrations. This highlights the practicality and effectiveness of our approach, particularly in strict zero-shot scenarios where annotated data is unavailable or expensive to obtain.

\subsection{Comparison with CoT methods}
\label{sec:appendix:additional_analyses:CoT}

\noindent In this subsection, we compare \abbr~with CoT-based methods and analyze the computational cost of different approaches on MMLU. 
We compare our \abbr~framework with single-step reasoning (Baseline), Chain of Thought (CoT, \cite{wei2022chain}), and Self-Consistency with Chain of Thought (CoT-SC, \cite{wang2022self}) in Table~\ref{table_CoT}. 
In the case of CoT, performance actually degraded across all benchmarks, and CoT-SC—which involves five reasoning passes—showed poor performance on EgoSchema, a dataset with long video contexts. This suggests that unconditionally incorporating unverified intermediate steps (such as sub-QAs) can not only underperform compared to simple single-step reasoning but also lead to inefficiencies.


\begin{table}[t]
\centering
\resizebox{\linewidth}{!}{
    \begin{tabular}{l|c|ccc}
    \toprule
    & \textbf{avg. \# path}\footnotemark[2] & \textbf{MMLU} & \textbf{MMMU} & \textbf{EgoSchema} \\
    \midrule
    
    Baseline                & 1   & 67.1 & 50.2 & 68.0 \\
    \midrule
    CoT                     & 1   & 65.3 & 49.8 & 62.3 \\
    CoT-SC                  & 5   & 68.3 & \underline{51.9} & 67.0 \\
    CoT-SC                  & 40  & \textbf{71.0} & \textbf{56.6} & \underline{69.0} \\
    \textbf{\abbr~(Ours)}   & 2.3 & \underline{69.2} & \underline{51.9} & \textbf{71.5} \\
    \textbf{\abbr~(Ours)$*$}& 1.7 & \underline{69.2} & 51.8 & \textbf{71.5} \\

    
    
    
    \bottomrule
    \end{tabular}
}
\caption{Comparison with CoT-based methods (Qwen2.5-VL backbone). 
The total number of reasoning paths is averaged over all main questions and reported as `avg. \# path.' 
Bold text indicates the highest score, while underlined text represents the second-highest score.  
}
\label{table_CoT}
\end{table}


    
    
    


\begin{table}[t]
\centering
\resizebox{\linewidth}{!}{
    \begin{tabular}{l|c|cc}
    \toprule
    \textbf{MMLU} & \textbf{avg. \# path}\footnotemark[2] & \textbf{Input / Generate tokens} & \textbf{Cost} \\ 
    \midrule
    Baseline                 & 1     & 132 / 13      & 1 \\ 
    \midrule
    
    CoT                      & 1     & 168 / 286     & 7.2 \\
    CoT-SC                   & 5     & 830 / 1427    & 35.7 \\
    \textbf{\abbr~(Ours)}    & 2.3   & 708 / 84      & 5.7 \\
    \textbf{\abbr~(Ours)$*$} & 1.7   & 442 / 52      & 3.5 \\
    
    \bottomrule
    \end{tabular}
}
\caption{
    Cost analysis of different approaches on MMLU. We represent the cost as a value normalized by the cost of the baseline (single-step reasoning), which is set to 1. `avg. \# path' denotes average of the total number of reasoning paths. 
}
\label{table_cost}
\end{table}
\begin{table*}[t]
    \resizebox{1.0\linewidth}{!}{
    \begin{tabular}{l|l|@{\hskip 6pt}lccccccccccccccc@{\hskip 6pt}}
        \toprule
        \multirow{2}{*}{\textbf{Backbone}} & \multirow{2}{*}{\textbf{Size}}& \multirow{2}{*}{\textbf{Reasoning Method}} &
        \multirow{2}{*}{\boldmath{$N$}} & \multirow{2}{*}{\boldmath{$M$}} & \multirow{2}{*}{\boldmath{$K$}} & 
        \multicolumn{3}{c}{\textbf{Text-only QA}} & \multicolumn{1}{c}{\textbf{ImageQA}} & \multicolumn{1}{c}{\textbf{VideoQA}}  \\
        \cmidrule(lr){7-9} \cmidrule(lr){10-10} \cmidrule(lr){11-11}
        & & & & & & \textbf{MMLU} & \textbf{MMLU-Pro} & \textbf{StrategyQA} & \textbf{MMMU} & \textbf{EgoSchema}  \\ \midrule

        \multirow{10}{*}{\makebox[0.3em][c]{\rotatebox{90}{Qwen2.5}}} & 7B & 
        Baseline                                 & - & - & - & 67.1          & 38.6          & 62.6          & 50.2          & 68.0          \\
        && \textsc{SingleSubQA}                  & 1 & 1 & 1 & \textcolor{blue}{67.5}   & \textcolor{blue}{42.9}   & 62.6          & \textcolor{red}{47.6}    & \textcolor{blue}{69.2}   \\
        && \textsc{EverySubQA}                   & 5 & 5 & 1 & \textcolor{blue}{67.7}   & \textcolor{blue}{42.6}   & \textcolor{blue}{\underline{65.3}} & \textcolor{red}{47.6}    & \textcolor{blue}{69.0}   \\
        && \subqajudge~   (sub-Q)                & 5 & 2 & 1 & \textcolor{blue}{67.5}   & \textcolor{blue}{42.7}   & \textcolor{blue}{64.0}   & \textcolor{red}{48.2}    & \textcolor{blue}{69.2}   \\
        && \subqajudge~   (sub-A)                & 5 & 2 & 1 & \textcolor{blue}{67.3}   & \textcolor{blue}{43.2}   & \textcolor{blue}{64.5}   & \textcolor{red}{49.4}    & \textcolor{red}{67.0}    \\
        && \subqajudge~   (sub-QA)               & 5 & 2 & 1 & \textcolor{blue}{67.2}   & \textcolor{blue}{42.8}   & \textcolor{blue}{64.1}   & \textcolor{red}{49.9}    & 68.0          \\
        && {\textbf{\abbr (Ours)} (seq. prob.)}   & 5 & 2 & 4 & \textcolor{blue}{\underline{68.1}} & \textcolor{blue}{39.8}   & \textcolor{blue}{63.5}   & \textcolor{blue}{\underline{51.2}} & \textcolor{blue}{\underline{70.2}} \\
        && {\textbf{\abbr (Ours)} ($1/$PPL)}       & 5 & 2 & 4 & \textcolor{blue}{67.8}   & \textcolor{blue}{\underline{43.4}} & \textcolor{blue}{63.6}   & \textcolor{blue}{50.7}   & \textcolor{blue}{\textbf{71.5}} \\
        && {\textbf{\abbr (Ours)} (min. token prob.)} & 5 & 2 & 4 & \textcolor{blue}{\textbf{69.2}} & \textcolor{blue}{\textbf{44.3}} & \textcolor{blue}{\textbf{65.7}} & \textcolor{blue}{\textbf{51.9}} & \textcolor{blue}{\textbf{71.5}} \\
        && \multicolumn{1}{l}{\blue{\grey{~$\Delta$}}} & \grey{} & \grey{} & \grey{} & \blue{\bf \grey{+2.1}} & \blue{\bf \grey{+5.7}} & \blue{\bf \grey{+3.1}}& \blue{\bf \grey{+1.7}} & \blue{\bf \grey{+3.5}} \\
        \midrule

        \multirow{10}{*}{\makebox[0.3em][c]{\rotatebox{90}{LLaVA-Onevision}}} & 7B & 
        Baseline                                 & - & - & - & 66.2          & 35.7          & 64.5          & 45.1          & 43.0          \\
        && \textsc{SingleSubQA}                  & 1 & 1 & 1 & \textcolor{red}{64.3}    & \textcolor{blue}{36.6}   & \textcolor{blue}{67.8}   & \textcolor{red}{44.1}    & \textcolor{red}{40.2}    \\
        && \textsc{EverySubQA}                   & 5 & 5 & 1 & \textcolor{red}{63.4}    & \textcolor{blue}{37.0}   & \textcolor{blue}{67.6}   & \textcolor{blue}{46.1}   & \textcolor{red}{39.8}    \\
        && \subqajudge~   (sub-Q)                & 5 & 2 & 1 & \textcolor{red}{64.9}    & \textcolor{blue}{37.5}   & \textcolor{blue}{67.5}   & \textcolor{blue}{46.4}   & \textcolor{blue}{44.8}   \\
        && \subqajudge~   (sub-A)                & 5 & 2 & 1 & \textcolor{red}{65.1}    & \textcolor{blue}{36.9}   & \textcolor{blue}{\underline{68.2}} & \textcolor{blue}{47.2}   & \textcolor{blue}{45.2}   \\
        && \subqajudge~   (sub-QA)               & 5 & 2 & 1 & \textcolor{red}{64.6}    & \textcolor{blue}{37.8}   & \textcolor{blue}{68.1}   & \textcolor{blue}{47.0}   & \textcolor{blue}{\underline{45.8}} \\
        && {\textbf{\abbr (Ours)} (seq. prob.)}   & 5 & 2 & 4 & \textcolor{blue}{\underline{67.1}} & \textcolor{blue}{\underline{38.0}} & \textcolor{blue}{66.7}   & \textcolor{blue}{\underline{47.3}} & \textcolor{blue}{43.5}   \\
        && {\textbf{\abbr (Ours)} ($1/$PPL)}       & 5 & 2 & 4 & 66.2          & \textcolor{blue}{37.6}   & \textcolor{blue}{67.8}   & \textcolor{blue}{\textbf{47.4}} & \textcolor{blue}{44.2}   \\
        && {\textbf{\abbr (Ours)} (min. token prob.)} & 5 & 2 & 4 & \textcolor{blue}{\textbf{67.3}} & \textcolor{blue}{\textbf{38.4}} & \textcolor{blue}{\textbf{68.3}} & \textcolor{blue}{\textbf{47.4}} & \textcolor{blue}{\textbf{46.0}} \\
        && \multicolumn{1}{l}{\blue{\grey{~$\Delta$}}} & \grey{} & \grey{} & \grey{} & \blue{\bf \grey{+1.1}} & \blue{\bf \grey{+2.7}} & \blue{\bf \grey{+3.8}}& \blue{\bf \grey{+2.3}} & \blue{\bf \grey{+3.0}}  \\

        \bottomrule
    \end{tabular}
}
    \caption{
    \textbf{Zero-shot evaluation results of \abbr.}
    We evaluate baseline models and our framework across various domains and benchmarks for comparison. $N$, $M$, and $K$ denote the number of generated sub-QAs, the number of curated sub-QAs per reasoning path, and the total number of reasoning paths, respectively.
    The difference (\blue{$\Delta$}) denotes the score improvement of \abbr~over the vanilla models without sub-QAs. 
    \textbf{Bold text} indicates the highest score and \underline{underlined text} the second-highest. Colors show performance relative to baseline: \textcolor{blue}{blue} (higher), \textcolor{red}{red} (lower).
    }
    \label{tab:additional_result}
\end{table*}

\footnotetext[2]{
Due to Principle 1 of \S~\ref{sec:methods:answer_selector}, if the confidence of the base answer for a main question is sufficiently high, all subsequent steps are skipped, resulting in a total of 1 reasoning path. In the case of MMLU, this applies to approximately 68\% of instances. When $K=4$, the average number of reasoning paths becomes $0.68 \times 1 + 0.32 \times (1 + 4) \simeq 2.3$.
}

\subsection{Cost analysis}
\label{sec:appendix:additional_analyses:cost}

To analyze the cost of \abbr, we report the average number of input and generated tokens in Table~\ref{table_cost}. The `Cost' column is represented relative to the baseline, which is normalized to 1, and the cost of generated tokens is counted as four times higher, following the pricing of models like GPT-4o. Compared to CoT-based methods, our approach achieves competitive performance (Table~\ref{table_CoT}) with lower cost. This is because \abbr~skips subsequent steps when the confidence of the single-step reasoning is high, avoiding complex computation for relatively simple questions. For MMLU, up to 68\% of the main questions are answered using only single-step reasoning (this value varies depending on the benchmark). These results support that \abbr~avoids unnecessary computation and performs inference efficiently.

Furthermore, we observe that if the confidence of a refined answer is sufficiently high, it is not necessary to explore all $K$ reasoning paths to maintain performance. For example, if the generation of refined answers is stopped when the confidence score exceeds 0.85, the overall cost can be reduced by 39\% while preserving nearly the same level of performance. This result is reported as `\abbr~(Ours)($*$)' in Tables~\ref{table_CoT} and~\ref{table_cost}. 

\subsection{Additional direct verification methods}
\label{sec:appendix:additional_analyses:subqa_judge}

We compare \abbr~with methods that use only sub-QAs verified through direct LLM prompting (\subqajudge), and present the results in Table~\ref{tab:additional_result}.
We experiment with three settings: verifying only the quality of sub-questions (sub-Q), the correctness of sub-answers (sub-A), and both (sub-QA). In all cases, performance falls short of our proposed framework and, in some instances, even underperforms compared to the baseline. These results highlight that simply relying on LLM prompting to validate intermediate steps is neither fully reliable nor consistently effective.
The prompt used for verifying sub-QAs can be found in Appendix~\ref{sec:appendix:prompt}.

\subsection{Discussion of confidence score metrics} 
\label{sec:appendix:additional_analyses:calibration}

\begin{figure}[t] 
  \centering
  \includegraphics[width=0.99\linewidth]{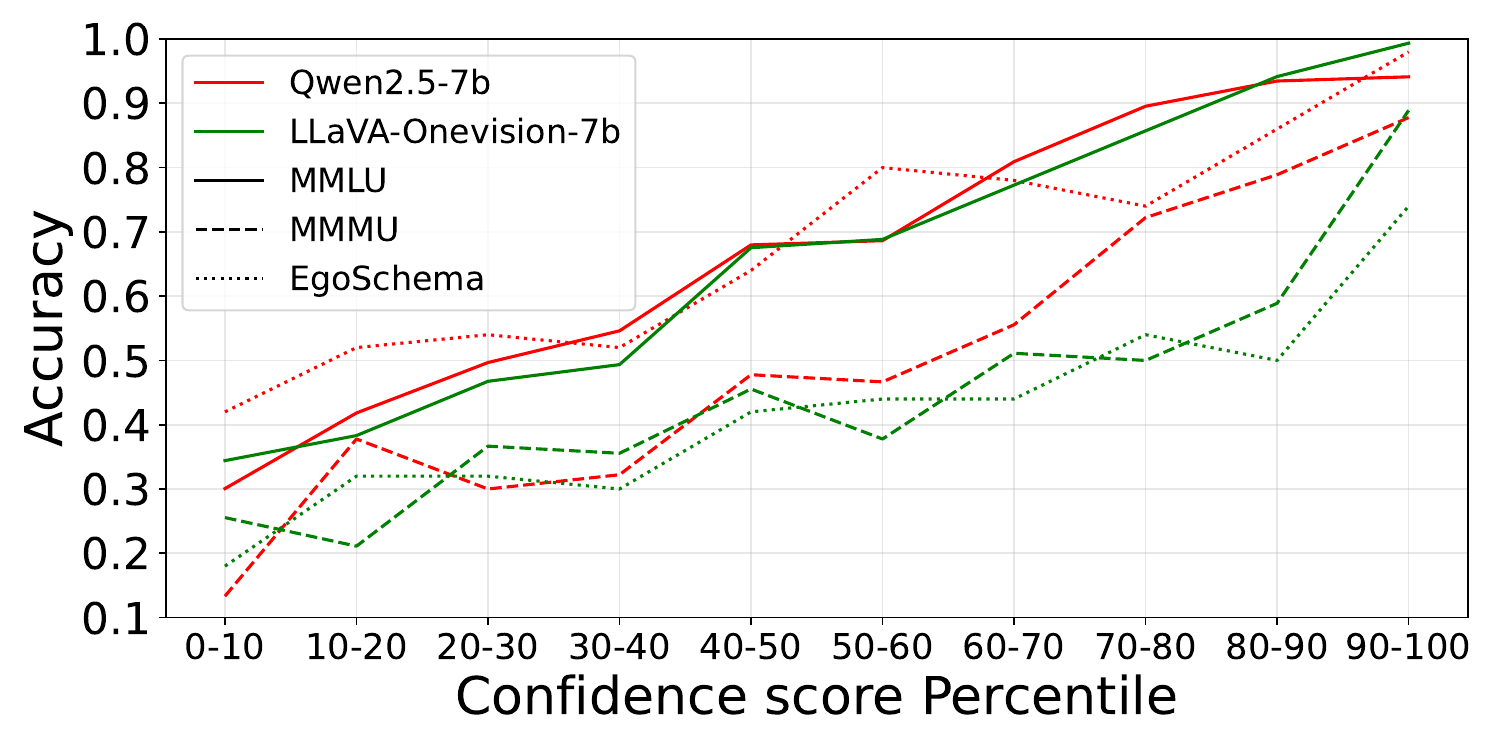}
  \caption{
    Correlation between confidence score and accuracy. We compute the base answer for each data instance, then sort these instances in ascending order of confidence and divide them into 10 bins. The results demonstrate a strong correlation between a response's confidence score and accuracy.
  }
  \label{fig:conf_and_acc}
\end{figure}

\begin{table}[t]
\centering
\resizebox{\linewidth}{!}{
    \begin{tabular}{lccc}
    \toprule
    \textbf{Method} & \textbf{MMLU} & \textbf{MMMU} & \textbf{EgoSchema} \\
    \midrule
    Baseline            & 67.1 & 50.2 & 68.0 \\
    \abbr~(Normalized)  & 68.9 & 51.4 & 71.2 \\
    \abbr~(Ours)        & \textbf{69.2} & \textbf{51.9} & \textbf{71.5} \\
    \bottomrule
    \end{tabular}
}
\caption{Performance comparison on Qwen2.5 using normalized confidence scores versus our proposed threshold-based method. While normalization improves upon the baseline, our approach remains superior.}
\label{tab:normalization_ablation}
\end{table}

We compare three different metrics for calculating the confidence score of a generated answer: 1) minimum token probability (\S\ref{sec:methods:preliminaries}); 2) sequence probability (i.e., the generation probability of the entire sequence); and 3) the reciprocal of perplexity (i.e., $1/\text{PPL}$).
As shown in Table~\ref{tab:additional_result}, using the minimum token probability consistently outperforms the other metrics in all cases. This supports the suitability of minimum token probability as a robust criterion for measuring confidence.

Also, for our framework to function effectively, the fundamental statement must hold that \textit{if a model assigns a high confidence score to an answer, the answer is more likely to be correct.} Since we use the minimum token probability of an answer as its confidence score, it is necessary to examine the correlation between this probability and the actual accuracy.
To validate this, we compute the base answers and their corresponding confidence scores (i.e., minimum token probability) for all data instances and sort them in ascending order based on confidence. We then divide the sorted instances into 10 bins and calculate the accuracy for each bin. As shown in Figure~\ref{fig:conf_and_acc}, accuracy increases with confidence. Also, the Pearson correlation coefficient, averaged across all models and benchmarks, is as high as 0.95. This confirms that the minimum token probability of an answer is a reliable metric for selecting the best answer.

We also tested an alternative method for comparing scores from base and refined answers: normalization. In an ablation study, we normalized both scores (zero mean, unit variance) and selected the one with the higher value, foregoing the threshold $\tau_2$.
Table~\ref{tab:normalization_ablation} shows that while normalization beats the baseline, our threshold-based method is superior. This validates our design, as it implies that simple normalization is insufficient to align the different confidence distributions of base and refined answers. Our threshold mechanism is therefore essential for handling these differences.

\section{Benchmarks}
\label{sec:appendix:benchmarks}

We evaluate models on five challenging QA benchmarks:

\noindent{\textbf{MMLU}}~\cite{hendrycks2020measuring} is a benchmark designed to measure a model's multitask accuracy across 57 diverse tasks like elementary mathematics, history, and computer science. It demands extensive world knowledge and problem-solving skills from models. The benchmark aims to analyze models broadly and identify their key shortcomings in academic and professional understanding. MMLU is MIT licensed.

\noindent{\textbf{MMLU-Pro}}~\cite{wang2024mmlu} is an enhanced and more challenging version of MMLU that integrates more reasoning-intensive questions and eliminates noisy ones. It spans 14 domains with over 12,000 questions and features ten answer choices to assess true understanding better. MMLU-Pro is designed to more effectively discriminate between advanced language models as their performance on the original MMLU has plateaued. MMLU-Pro is MIT licensed.

\noindent{\textbf{StrategyQA}}~\cite{geva2021didaristotleuselaptop} is a QA dataset where the reasoning steps needed to answer questions are implicit and must be inferred. Its data collection process elicits creative questions and includes adversarial filtering. Each of its 2,780 examples provides the question, a reasoning decomposition, and supporting evidence paragraphs. StrategyQA is MIT licensed.

\noindent{\textbf{MMMU}}~\cite{yue2024mmmumassivemultidisciplinemultimodal} is a benchmark with 11,500 multimodal questions from college exams and textbooks across six core disciplines like Science, Engineering, and Art \& Design. It assesses expert-level multimodal understanding, requiring college-level knowledge and reasoning with diverse image types like charts and diagrams. MMMU challenges models on advanced perception and domain-specific reasoning. MMMU is Apache License 2.0 licensed.

\noindent{\textbf{EgoSchema}}~\cite{mangalam2023egoschema} is a very long-form video question-answering dataset from Ego4D, designed to evaluate long video understanding. It includes over 5,000 human-curated multiple-choice questions on more than 250 hours of real-world video depicting natural human activities. EgoSchema specifically tests a model's ability to comprehend extended temporal structures and complex behaviors in video. EgoSchema is open-sourced under the Ego4D license.

\section{Details of Models}
\label{sec:appendix:model_details}

In this section, we describe three vision-language foundation models, each based on a different LLM, that are used in our experiments: Qwen2.5-VL~\cite{yang2024qwen2}, Gemma 3~\cite{team2025gemma}, and LLaVA-Onevision~\cite{li2024llavaonevisioneasyvisualtask}. 
We use the checkpoint provided by huggingface.

\noindent \textbf{Qwen2.5-VL} is a vision-language model series based on Qwen2.5 LLM, showcasing significant advancements in visual understanding, object localization, and long-video comprehension. It introduces dynamic resolution processing and absolute time encoding, enabling efficient handling of images with varying sizes and videos spanning several hours. Also, Qwen2.5 is a versatile series of large language models significantly improved through expanded pre-training (18 trillion tokens) and advanced post-training techniques, including supervised finetuning and reinforcement learning. These enhancements lead to strong performance in reasoning, instruction following, long-text generation, and structured data analysis.

\noindent \textbf{Gemma 3} is a multimodal extension of the Gemma model family, available in sizes from 1B to 27B parameters, with added vision capabilities and support for long contexts up to 128K tokens. Its architecture is optimized for memory efficiency by increasing the proportion of local attention layers and shortening local attention spans. A new post-training method and distillation enhance its performance in math, chat, multilingual understanding, and instruction following. All models are openly released to the community.

\noindent \textbf{LLaVA-Onevision} is a family of open large multimodal models designed to unify insights from data, model architecture, and visual representations. It is the first single model to advance performance across single-image, multi-image, and video scenarios simultaneously. The model also exhibits strong transfer learning across modalities, enabling emergent capabilities such as video understanding through image-to-video task transfer.

\section{Details of Experiment Settings}
\label{sec:appendix:settings}
When evaluating on multiple-choice QA benchmarks, we include all answer choices in the input and have the model select the correct one.
For \VQA, we sample each video at 1fps; for videos over 32 seconds, we uniformly sample 32 frames from the entire video.
We report performance on the validation set based on a single run; however, for datasets without a standard validation split (StrategyQA) or with very few validation examples, we evaluate on another split instead. For MMLU-Pro, since the validation set contains only 70 examples, we use the test set instead.
As of May 2025, submissions to the StrategyQA leaderboard are impossible, so we report results on the training set.
We tuned the confidence thresholds, $\tau_1$ and $\tau_2$, as hyperparameters on a validation set for each benchmark. The optimal values were identified through a grid search with a step size of 0.1. The search range for $\tau_1$, which is compared against a confidence probability, was $[0, 1]$. For $\tau_2$, representing the difference between two probabilities, the range was $[-1, 1]$. For datasets without an official validation split, we held out 10\% of the training data for this purpose.
For the closed-source models (i.e., GPT-4o), we enable the `logprobs' option to obtain the probability of each generated token and use the `gpt-4o-2024-08-06' version.

\section{Prompt Designs}
\label{sec:appendix:prompt}

\begin{figure*}[!ht]
  \centering
  \includegraphics[width=0.75\textwidth]{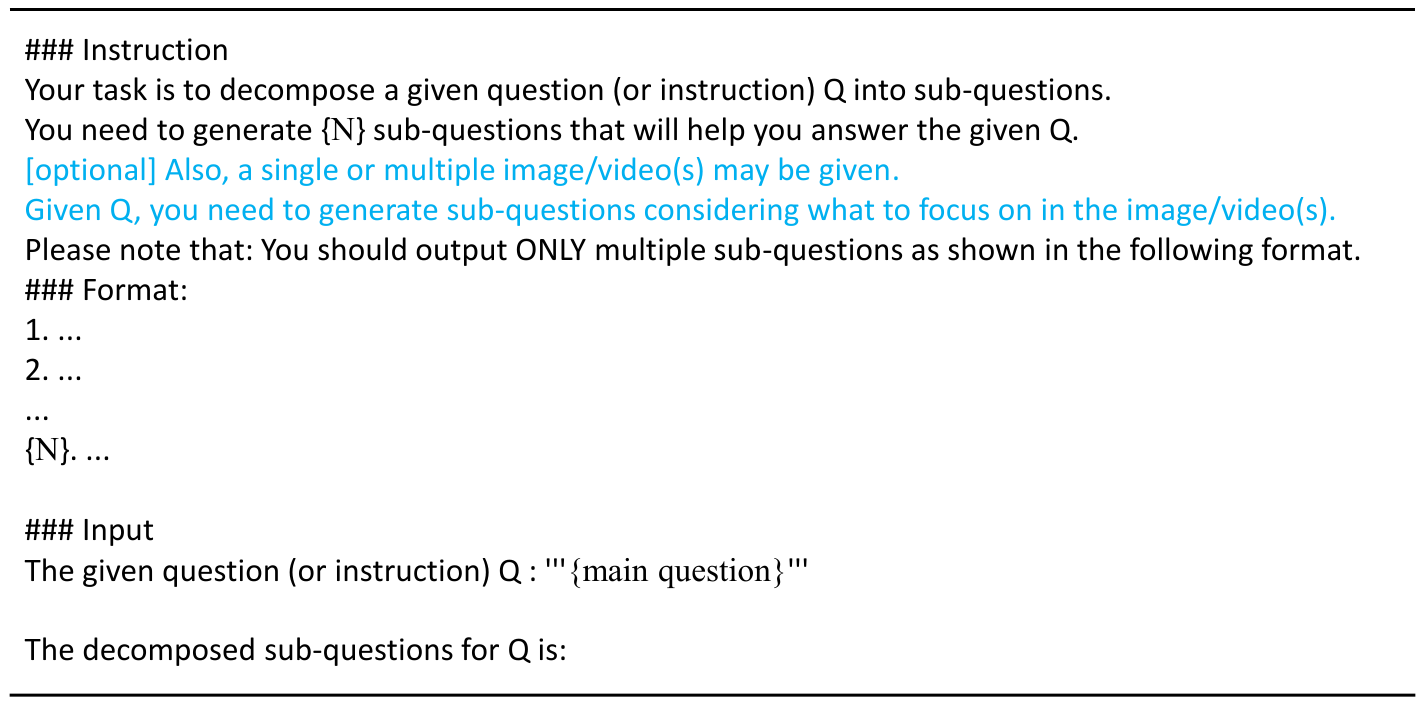}
  \caption{
    The prompt for generating sub-questions. Blue text is excluded from the prompt when performing Text-only QA. \{N\} denotes the number of sub-questions to be generated.
  }
  \label{fig:prompt_subq}
\end{figure*}
\begin{figure*}[!ht]
  \centering
  \includegraphics[width=0.99\textwidth]{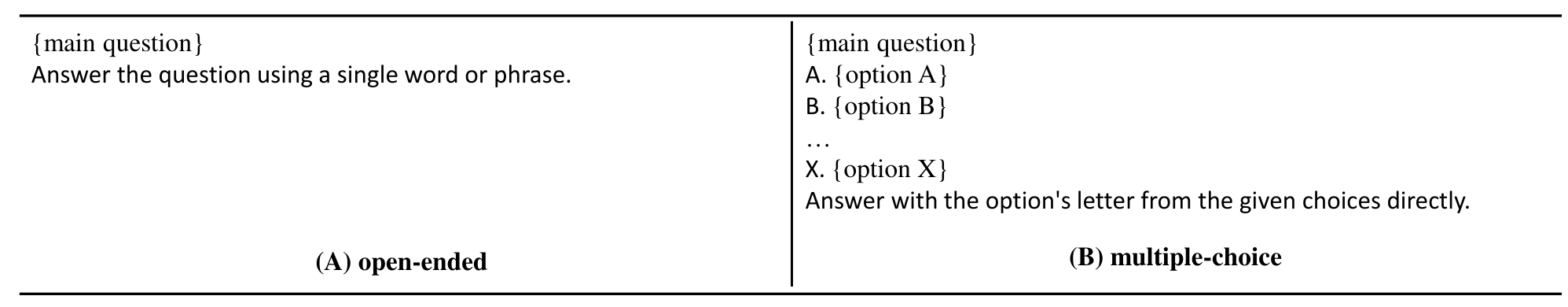}
  \caption{
    The prompt for generating base answer $A_\text{base}$. \textbf{(A)} corresponds to open-ended questions, while \textbf{(B)} represents multiple-choice questions. Option X refers to the last option, with the total number of options varying depending on the question.
  }
  \label{fig:prompt_base}
\end{figure*}
\begin{figure*}[!ht]
  \centering
  \includegraphics[width=0.99\textwidth]{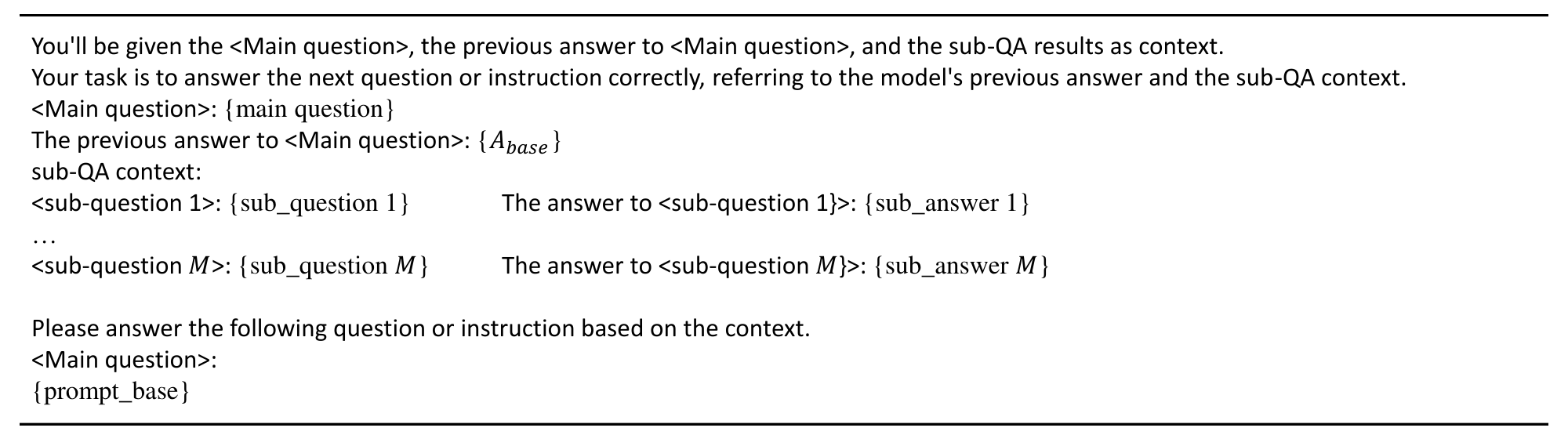}
  \caption{
    The prompt for generating refined answer $\hat{A}_{\text{refined}}$. The placeholder `\{prompt\_base\}' (referring to the prompt used to generate the base answer) is detailed in Figure~\ref{fig:prompt_base}.
  }
  \label{fig:prompt_refined}
\end{figure*}
\begin{figure*}[!ht]
  \centering
  \includegraphics[width=0.99\textwidth]{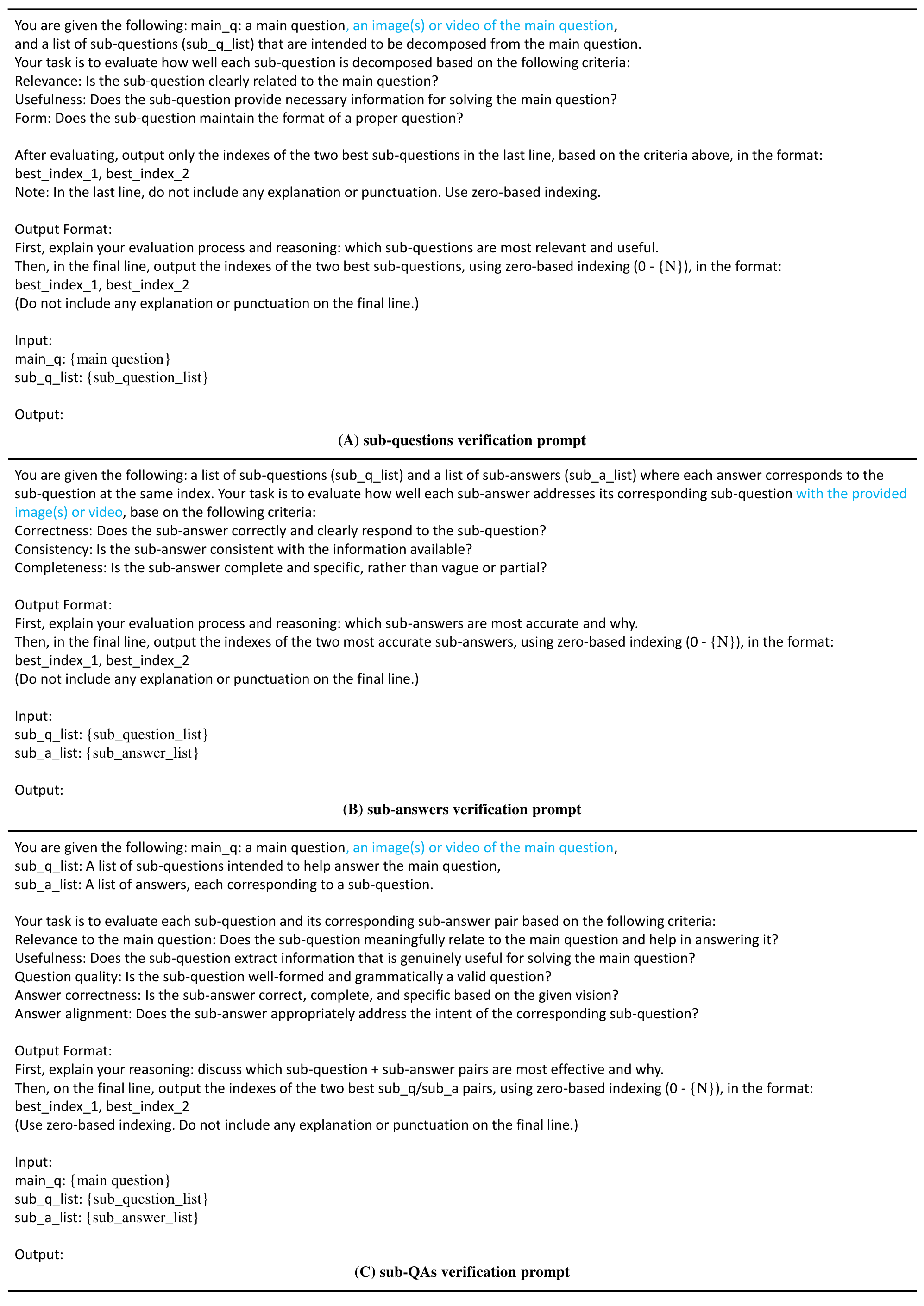}
  \caption{
    The prompt for verifying sub-questions and sub-answers directly (\subqajudge). These sub-QA verification prompts may not be optimal.
  }
  \label{fig:prompt_subqajudge}
\end{figure*}

We briefly describe the prompts used for generating sub-QAs, base answers, and refined answers:

\begin{itemize}
    \item We use the prompt shown in Figure~\ref{fig:prompt_subq} when generating sub-questions.
    \item For obtaining sub-answers, we use the following simple prompt: ``\{sub-question\} Answer in a maximum of one sentence.''
    \item We design prompts to generate base answers for open-ended and multiple-choice QA tasks, as shown in Figure~\ref{fig:prompt_base}. For open-ended QA, we use ``Question: \{main question\} Answer the question using a single word or phrase.'' 
    For multiple-choice QA, the options are added after the main question: ``A. \{option A\} B. \{option B\} \dots~X. \{option X\}.'' Option X refers to the last option, with the total number of options varying depending on the question.
    \item For refined answers, we use the prompt shown in Figure~\ref{fig:prompt_refined}.
    \item The prompt used to verify sub-QAs in the \subqajudge~method is presented in Figure~\ref{fig:prompt_subqajudge}.
\end{itemize}

\section{Failure Cases in sub-QA Generation}
\label{sec:appendix:failure_analysis}

While our \abbr~framework demonstrates strong performance, its effectiveness is inherently dependent on the quality of the initial sub-QA generation. The framework's limitations become apparent when this process yields incomplete or flawed information, as we analyze in this section.

The primary failure modes revealed by our analysis are:
\begin{itemize}
    \item \textbf{Irrelevant or semantically drifting sub-questions:} Sub-questions \#2 and \#4 are only tangentially related to the core reasoning path. They define ``mascot'' in a general sense or explore irrelevant aspects of a mascot's ``diet,'' which does not help solve the specific question.
    \item \textbf{Incomplete sub-answers:} Sub-question \#1 is relevant, but its answer crucially omits the key information needed for reasoning—the name of the high school Jindal attended.
    \item \textbf{Redundancy and semantic overlap:} Sub-questions \#4 and \#5 are semantically similar. While we note that exact duplication of sub-questions is rare (occurring in less than 0.1\% of cases), this kind of semantic overlap can lead to inefficient reasoning.
\end{itemize}
In the example shown, only sub-QA \#3 provides a directly useful fact. Due to the lack of critical connecting information from other sub-QAs, the Refiner failed to construct a correct reasoning path, ultimately leading to an incorrect output in most refined candidates.

\begin{table*}[t]
\centering
\caption{A representative failure case from the StrategyQA dataset. The model fails because the generated sub-questions do not furnish the critical information needed for the required multi-hop reasoning.}
\label{tab:failure_case}
\begin{tabular}{l p{0.80\textwidth}}
\toprule
\textbf{Main Question} & Would Bobby Jindal's high school mascot eat kibble? \\
\textbf{Ground Truth}  & Yes. (\textit{Reasoning: Bobby Jindal attended Baton Rouge Magnet High, whose mascot is a bulldog. Bulldogs are dogs, and kibble is dog food.}) \\
\textbf{Base Answer}   & Yes. \\
\midrule
\multicolumn{2}{c}{\textbf{Generated Sub-Question and Sub-Answer Pairs}} \\
\midrule
\textbf{Sub-Q \#1} & Who is Bobby Jindal? \\
\textbf{Sub-A \#1} & \textit{Bobby Jindal was an American politician who served as the 47th Governor of Louisiana from 2008 to 2015 and was a Republican candidate for the 2016 presidential election.} \\
\midrule
\textbf{Sub-Q \#2} & What is a high school mascot? \\
\textbf{Sub-A \#2} & \textit{A high school mascot is a symbol or character representing a school, often used in sports teams and school spirit events.} \\
\midrule
\textbf{Sub-Q \#3} & What is kibble? \\
\textbf{Sub-A \#3} & \textit{Kibble is a dry, processed food for pets, typically made from a mixture of grains, proteins, and other ingredients, designed to be easily digestible and convenient for feeding.} \\
\midrule
\textbf{Sub-Q \#4} & What is the diet of a high school mascot? \\
\textbf{Sub-A \#4} & \textit{A high school mascot typically does not have a specific diet as they are not living organisms and do not consume food.} \\
\midrule
\textbf{Sub-Q \#5} & How do high school mascots typically eat? \\
\textbf{Sub-A \#5} & \textit{High school mascots typically eat like regular students, with meals and snacks similar to those consumed by their peers.} \\
\bottomrule
\end{tabular}
\end{table*}

\end{document}